\documentclass[runningheads]{llncs}
\usepackage{makeidx}
\usepackage{amsmath}
\usepackage{amsfonts}
\usepackage{amssymb}
\usepackage[numbers,sort]{natbib}
\usepackage{eucal}
\usepackage{bbold}  
\usepackage{algorithmicx,algorithm}
\usepackage[noend]{algpseudocode}
\usepackage{float}  
\usepackage{lipsum}
\makeatletter
\newenvironment{breakablealgorithm}
{
	\begin{center}
		\refstepcounter{algorithm}
		\hrule height.8pt depth0pt \kern2pt
		\renewcommand{\caption}[2][\relax]{
			{\raggedright\textbf{\ALG@name~\thealgorithm} ##2\par}%
			\ifx\relax##1\relax 
			\addcontentsline{loa}{algorithm}{\protect\numberline{\thealgorithm}##2}%
			\else 
			\addcontentsline{loa}{algorithm}{\protect\numberline{\thealgorithm}##1}%
			\fi
			\kern2pt\hrule\kern2pt
		}
	}{
		\kern2pt\hrule\relax
	\end{center}
}
\makeatother
\usepackage{graphicx}
\makeindex

\begin{document}
\title{Generative Face Parsing Map Guided 3D Face Reconstruction Under Occluded Scenes}
%
%
 \author{Dapeng Zhao\inst{1} \and
 Yue Qi\inst{1,2,3} }
\institute{State Key Laboratory of Virtual Reality Technology and Systems,School of Computer Science and Engineering at Beihang University, Beijing, China  \\
 \email{mirror1775@gmail.com}\\
  \and
  Peng Cheng Laboratory, Shenzhen, China\\
  \and
  Qingdao Research Institute of Beihang University, Qingdao, China \\
  \email{}
 }
 \maketitle              
\ 
\\
\ 
\\
\begin{abstract}
    Over the past few years, single-view 3D face reconstruction methods can produce beautiful 3D models. Nevertheless, 
    the input of these works is unobstructed faces. 
    We describe a system designed to reconstruct convincing 
    face texture in the case of occlusion. 
    Motivated by parsing facial features, 
    we propose a complete face parsing map generation 
    method guided by landmarks. 
    We estimate the 2D face structure of the reasonable 
    position of the occlusion area, 
    which is used for the construction of 3D texture. 
    An excellent anti-occlusion face reconstruction 
    method should ensure the authenticity of the output, 
    including the topological structure between the eyes, 
    nose, and mouth. We extensively tested our method and 
    its components, qualitatively demonstrating the rationality 
    of our estimated facial structure. We conduct extensive 
    experiments on general 3D face reconstruction tasks as 
    concrete examples to demonstrate the method's superior 
    regulation ability over existing methods often break down. 
    We further provide numerous quantitative examples showing 
    that our method advances both the quality and the
     robustness of 3D face reconstruction under occlusion scenes.

\keywords{3D Face Reconstruction  \and Face Parsing \and Occluded Scenes.}
\end{abstract}

\section{Introduction}
3D face reconstruction refers to synthesizing a 
3D face model given one input face photo. 
It has a wide range of applications, such as face recognition 
and digital entertainment~\cite{wang2020leveraging}. Existing methods mainly 
concentrate on unobstructed faces, thus limiting the 
scenarios of their actual applications. Reconstructing 
a 3D face model from a single photo is a classical and 
fundamental problem in computer vision. 
The reconstruction task is challenging as human face 
structure partial invisibility when considering occluded scenes. 
Over the past five years, the related problem of face 
inpainting in images has gradually developed to the rationality 
of face photo generation in the most extreme 
scenes~\cite{RN578}. 
 \\
We cannot use artificial intelligence to robustly predict the 3D texture of the occluded area of the face. On the other hand, when faces are partially occluded, existing methods often indiscriminately reconstruct the occluded area. 
With the assistance of face parsing map,
 we find a way to identify the occluded area 
 and reconstruct the input image to a reasonable 3D face model.
The main contributions are summarized as follows: 
\\$\bullet$\ We propose a novel algorithm that combines feature points and 
face parsing map
 to generate face with complete facial features.
\\$\bullet$\ To address the problem of invisible face area under occluded scenes, we propose synthesizing input face photo based on Generative Adversarial Network rather than reconstructing 3D face directly.
\\$\bullet$\ We have improved the loss function of our 3D 
reconstruction framework for occluded scenes. 
Our method obtains state-of-the-art qualitative performance in real-world images. 
\section{Related Works}
\subsection{Generic Face Reconstruction}
The classic methods use reference 3D face models to fit the input face photo. 
Some recent techniques use Convolution Neural 
Networks (CNNs) to regress landmark locations with the 
raw face image.
 Some recent techniques firstly used CNNs to predict the 3DMM parameters with input face 
image. 
Some works proposed a cascaded CNN structure to regress the accurate 3DMM shape 
parameters~\cite{RN581,RN121,RN18,RN106,RN582}. Some frameworks explored the end-to-end CNN architectures to regress 3DMM coefficients directly. Each calculation usually takes a long time because the dimensionality of the data is very 
high~\cite{RN400}. 
\subsection{Face Parsing}A face parsing map generally serves as an intermediate representation for conditional face image 
generation~\cite{RN593}.
 In addition, the image-to-image GAN model can learn the mapping from the semantic map to realistic 
RGB image~\cite{RN257,RN594,RN327,RN390}[32-35]. In the pixel-level image semantic segmentation methods based on deep learning, 
FCN~\cite{RN613} is the well-known baseline for generic images which analyze per-pixel feature. Following this work, the DeepLab 
approaches~\cite{RN604,RN605,RN606,RN607} have achieved impressive results. The main feature of the series is to use hole convolution instead of traditional convolution. However, directly applying these frameworks for face parsing may fail to map the varying-yet-concentrated facial features, especially hair, leading to poor results. A workable solution should directly predict per-pixel semantic label across the entire face photo. 
Wei et al.~\cite{RN615} proposed a novel method for regulating receptive fields with superior regulation ability in parsing networks to access accurate parsing map. 
MaskGan~\cite{RN311} contributed a labeled face dataset~\cite{RN622}. 
Zhou et al.~\cite{RN616} proposed an architecture that explored how to combine the fully-convolutional network model and super-pixel data to model together. In order to solve the question of global image information access restriction, 
some methods~\cite{RN621,RN620,RN623,RN624} have introduced the transformer component and achieved state-of-the-art results. 
The semantic layout guides the location and appearance of facial features and further facilitates the training. The majority of face parsing methods work require semantic labels. Hence, these 
frameworks~\cite{RN311,RN48,RN600,RN569,RN601,RN37,RN79} usually train on the CelebA and Helen dataset which contain labeled attributes.
\subsection{Generative Adversarial Networks}Generative Adversarial Network 
(GAN)~\cite{RN744} generally consists of a generator and a discriminator. The two components compete with each other. Since GANs can generate realistic images, 
GANs have been successfully applied to various face image 
synthesis tasks, such as image manipulation~\cite{RN311,RN745,RN572,RN747},image-to-image 
translation~\cite{RN257,RN327,RN390,RN574,RN565}, image 
inpainting~\cite{RN751,RN753} and texture blend~\cite{RN748,RN749,RN750}.
For example, the face images generated by 
Stylegan2~\cite{RN754} can reach a high degree of recognition. 
With continuous improvements in 
regularization~\cite{RN327,RN573,RN677,RN678,RN673,RN316,RN697}, users can control 
the synthesis by feeding the generator with 
conditioning information instead of noise. 
Our work was built on conditional GANs~\cite{RN674} with 
face parsing map inputs, which aims to tackle 
face reconstruction under occluded scenes.
\subsection{Face Image Synthesis}Deep pixel-level 
face generating has been studied for 
a few years. Many methods achieve 
remarkable results. 
Context encoder~\cite{RN290} is the first 
deep learning network designed for image 
inpainting with encoder-decoder architecture. 
Nevertheless, the networks do a poor job in dealing 
with human faces. Following this work, 
Yang et al.~\cite{RN390} used a modified VGG 
network~\cite{RN617} to improve the result of the 
context-encoder, by minimizing the feature difference 
of photo background. Dolhansky et al.~\cite{RN618} 
demonstrated the significance of exemplar data 
for inpainting. However, this method only focuses on 
filling in missing eye regions of the frontal face, so 
it does not generalize well. 
EdgeConnect~\cite{RN353} shows 
impressive proceeds which disentangling generation 
into two stages: edge generator and image completion 
network. Contextual Attention~\cite{RN619} takes a similar 
two-step approach. First, it produces a base estimate of 
the invisible region. Next, the refinement block 
sharpens the photo by background patch sets. The 
typical limitations of current face image generate 
schemes are the necessity of manipulation, the complexity 
of fundamental architectures, the degradation in accuracy, 
and the inability of restricting modification to local region.
\section{Our Approach}
\begin{figure}[h]
\centering
\includegraphics[width=0.80\textwidth]{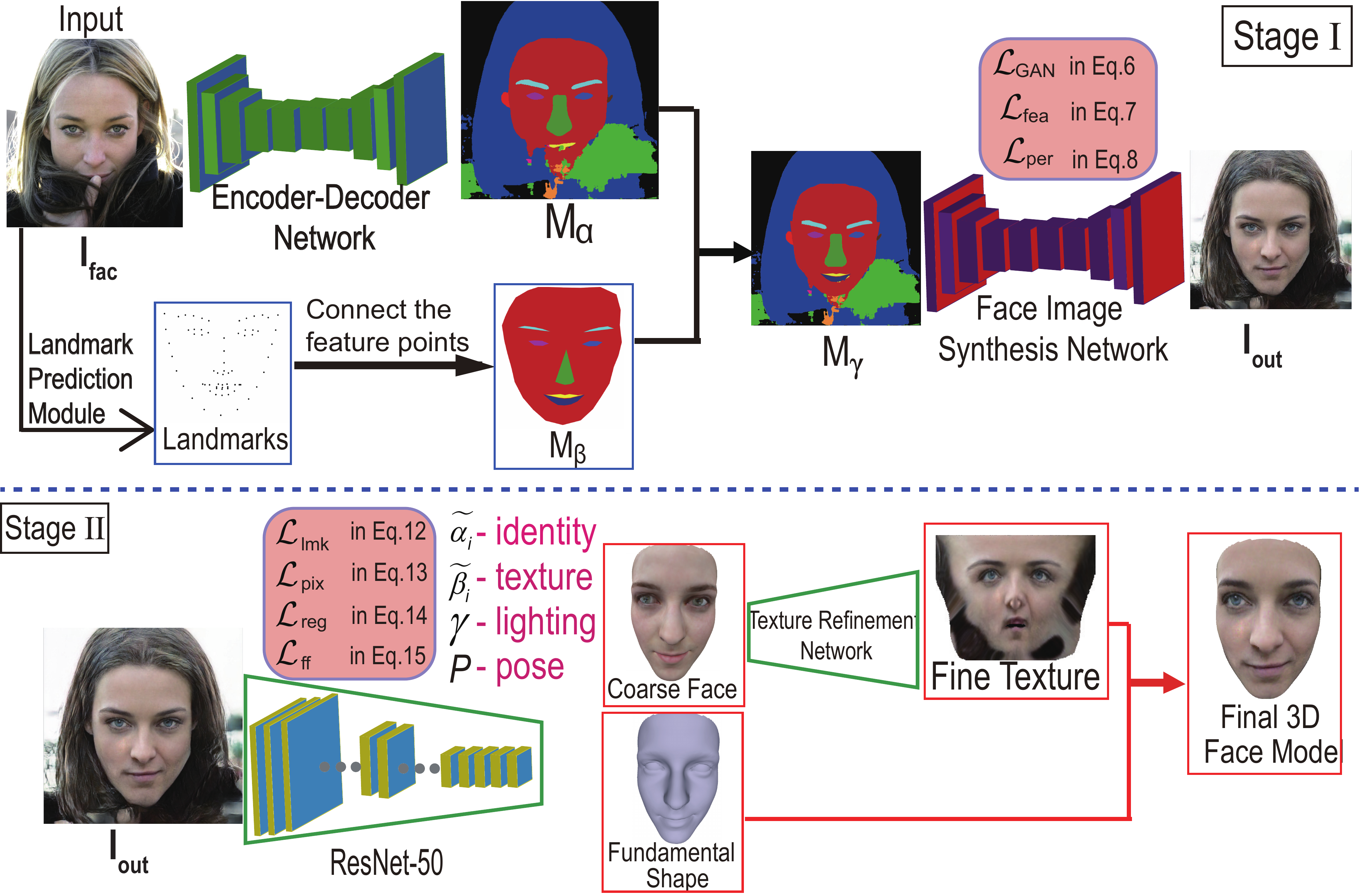}
\caption{Overall our pipeline. 
We first remove the occluded area and 
reconstruct the face with complete facial features. 
Then we utilize ResNet-50 and texture refinement 
network to reconstruct the final 3D model.} \label{fig:overall}
\end{figure}
\subsection{Landmark Prediction Task}
Fig.\ref{fig:overall} shows the entire process of our work.In the landmark prediction task, 
we found that generating accurate $68$ feature points 
${\mathbf{{Z}}_{\mathbf{lmk}}}\in {{\mathbb{R}}^{2\times 68}}$ was a crucial part under occlusion 
scenes. The 
architecture ${{\mathcal{N}}_\text{lmk}}$  aims to generate landmarks  from a corrupted face photo
${\text{I}_\text{cor}}:{\mathbf{Z}_\mathbf{lmk}}{=}{{\mathcal{N}}_{\text{lmk}}}\left( {{\text{I}}_{\text{cor}}};{{\theta }_{lmk}} \right)$ , 
where  ${{\theta }_{lmk}}$ denotes the trainable parameters.
 Since we want to focus more on efficiency and 
 follow face parsing map generation task, 
 we built a sufficiently effective ${{\mathcal{N}}_\text{lmk}}$  upon the 
 MobileNet-V3~\cite{RN740}. ${{\mathcal{N}}_\text{lmk}}$ is focused on feature extraction, unlike traditional landmark 
 detectors. The final module is realized by fully connecting the fused feature maps. 
We set the loss 
function ${{\mathcal{L}}_{lmk}}$ as follows:
\begin{equation}
{{\mathcal{L}}_{lmk}}\text{=}\left\| \mathbf{{Z}_{_{lmk}}^{(i)}}-\mathbf{{\hat{Z}}_{_{gt}}^{(i)}} \right\|_{2}^{2}
\end{equation}
    where $\mathbf{{\hat{Z}}_{_{gt}}^{(i)}}$  denotes the $i$th ground truth face landmarks.
\subsection{Face Parsing Map Generation}
\begin{figure}[htb]
    \centering
\includegraphics[width=0.50\textwidth]{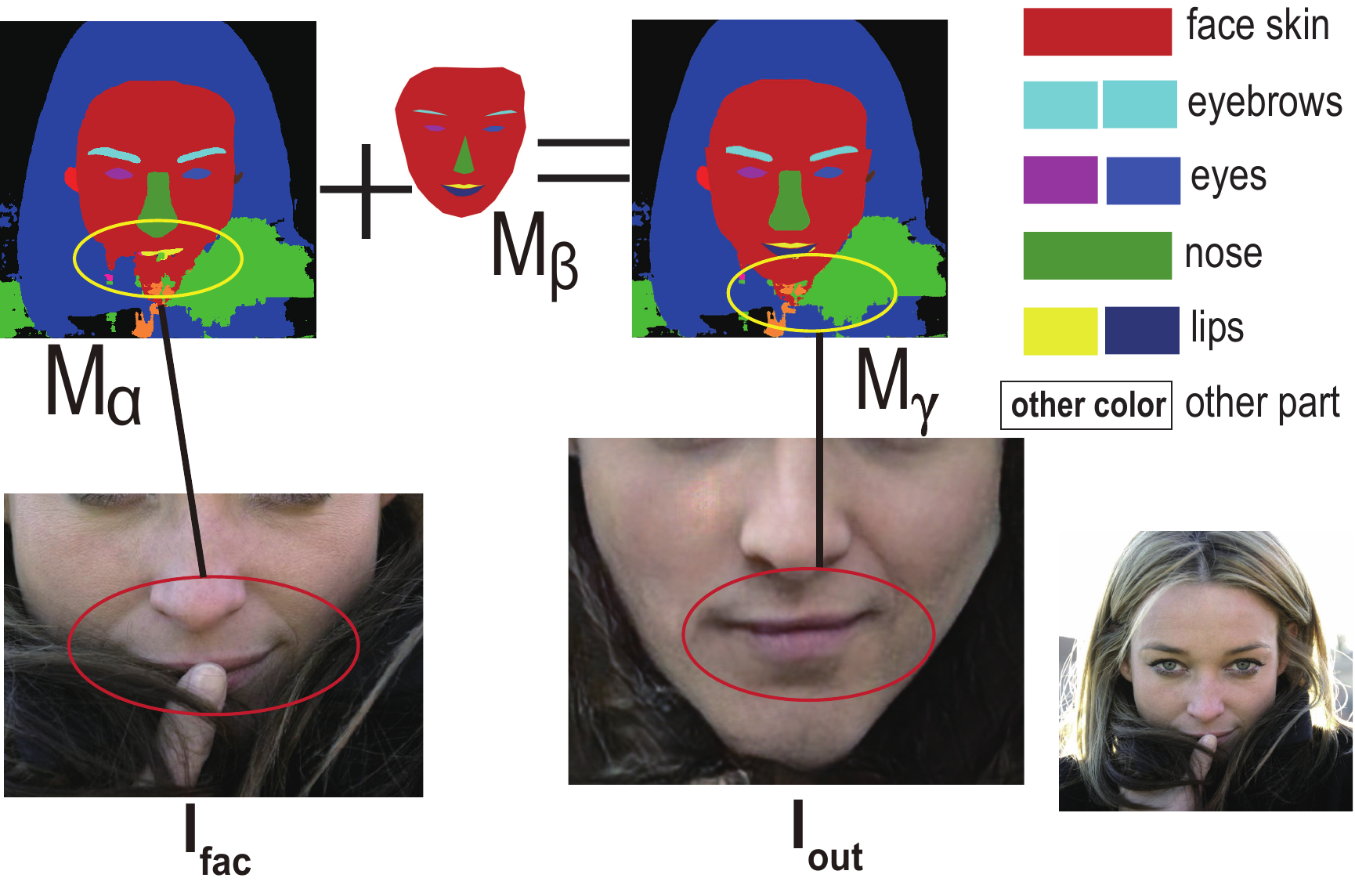}
\caption{Our face parsing map generation module, which follows
 Algorithm \ref{suanfa:001}. 
 The results shown in the figure show that 
 our method finally successfully removed 
 the occlusion of fingers and hair} \label{fig:plus}
\end{figure}
Pixel-level recognition 
of occlusion and face skin 
areas is a prerequisite for 
our framework to ensure accuracy. 
To benefit from the annotated face 
dataset CelebAMask-HQ~\cite{RN311}, we used 
an encoder-decoder architecture ${{\mathcal{N}}_{\alpha }}$   
based on U-Net~\cite{RN50} to estimate pixel-level 
label classes. Given a squarely resized face 
image ${{\mathbf{I}}_\mathbf{fac}}\in {{\mathbb{R}}^{H\times W\times 3}}$ , 
we applied the trained face parsing 
model ${{\mathcal{N}}_{\alpha }}$  to obtain the parsing 
map ${\mathbf{M}_{\boldsymbol{\alpha}}}\in {{\mathbb{R}}^{H\times W\times 1}}$ . 
On the other hand, given the 
landmarks ${\mathbf{Z}_\mathbf{lmk}}\in {{\mathbb{R}}^{2\times 68}}$, 
we connected the feature points to form 
a region. Then these regions can form a 
parsing map ${\mathbf{M}_{\boldsymbol{\beta}} } \in {\mathbb{R}^{H \times W \times 1}}$ including facial features. 
Please notice that, in our work, 
we assumed that facial features 
only include only five parts, 
including facial skin, eyebrows, 
eyes, nose and lips. 
The final map ${\mathbf{M}_{\boldsymbol{\gamma}} } \in {\mathbb{R}^{H \times W \times 1}}$   
(see Fig. \ref{fig:plus}) without 
occluded objects needs ${\mathbf{M}_{\boldsymbol{\alpha}} }$  plus ${\mathbf{M}_{\boldsymbol{\beta}} }$ . 
In order to generate ${\mathbf{M}_{\boldsymbol{\gamma}} }$  including 
the complete facial features, 
we designed Algorithm~\ref{suanfa:001}.
\begin{breakablealgorithm}
    \caption{Face Parsing Map Plus Algorithm, our 
    proposed algorithm. All experiments in the 
    papers Map $\mathbf{A}$ and Map $\mathbf{B}$ have the same width and height.}
    \label{suanfa:001}
     {\bf Input:}  ${{A}_{i}}$, pixels point on 
        the face parsing map $\mathbf{A}$. ${{B}_{i}}$, 
        pixels point on the face parsing map $\mathbf{B}$.
        $V\left(z \right)$, the function of getting 
        the grayscale value of point $z$.
        $X(i)$, the horizontal coordinate 
        value of $i$ in the map.
        $Y( i )$, the vertical 
        coordinate value of $i$  in the map. 
        $\text{W}$, the width of the map. $\text{H}$, the height 
        of the map. $\text{S}$, the gray value 
        range of the facial features area 
        (only include four parts:eyebrows, eyes, nose, lips).
         $\text{O}$, gray value range of the facial skin area.
         \\
    \leftline{\hspace*{0.02in} {\bf Input:} Face parsing map $\mathbf{A}$ and $\mathbf{B}$}

    \leftline{\hspace*{0.02in} {\bf Output:} ${{{C}}_{i}}$, pixels 
    point on the new face parsing map $\mathbf{C}$}
    \begin{algorithmic}[1]
\\{\bf while\ }{$Y(i)<=\text{H}$}{\ \bf do\ }       \Comment{Start to generate complete face skin}
\\\qquad\qquad{\bf while\ }{$X(i)<=\text{W}$}{\ \bf do\ }
\\\qquad\qquad\qquad{\bf if\ }${{A}_{i}}\in \text{O}$\ {\bf then\ }
\\\qquad\qquad\qquad\qquad${{C}_{i}}\gets {{A}_{i}},X( i )+1\gets X( i )$
\\\qquad\qquad\qquad{\bf else if\ }${{A}_{i}}\ ${\bf NOT\ }            $\in \text{O}$   {\bf AND}     ${{B}_{i}}\in \text{O}$          \ {\bf then\ }
\\\qquad\qquad\qquad\qquad${{C}_{i}}\gets {{B}_{i}},X( i )+1\gets X( i )$
\\\qquad\qquad\qquad{\bf else\ }${{C}_{i}}\gets {{A}_{i}},X( i )+1\gets X( i )$
\\\qquad\qquad{\bf end while\ }
\\\qquad$Y( i )+1\gets Y( i )$
\\{\bf end while\ }
\\
\\{\bf while\ }{$Y(i)<=\text{H}$}{\ \bf do\ }\Comment{Start to generate complete facial features}
\\\qquad\qquad{\bf while\ }{$X(i)<=\text{W}$}{\ \bf do\ }
\\\qquad\qquad\qquad{\bf if\ }${{A}_{i}}\in \text{S}$\ {\bf then\ }
\\\qquad\qquad\qquad\qquad${{C}_{i}}\gets {{A}_{i}},X( i )+1\gets X( i )$
\\\qquad\qquad\qquad{\bf else if\ }${{A}_{i}}\ ${\bf NOT\ }            $\in \text{S}$   {\bf AND}     ${{B}_{i}}\in \text{S}$          \ {\bf then\ }
\\\qquad\qquad\qquad\qquad${{C}_{i}}\gets {{B}_{i}},X( i )+1\gets X( i )$
\\\qquad\qquad\qquad{\bf else\ }${{C}_{i}}\gets {{A}_{i}},X( i )+1\gets X( i )$
\\\qquad\qquad{\bf end while\ }
\\\qquad$Y( i )+1\gets Y( i )$
\\{\bf end while\ }

    \end{algorithmic}
\end{breakablealgorithm}
\subsection{Face Image Synthesis with GAN}
\subsubsection{Face Image Synthesis Network}
To benefit from the Pix2Pix
architecture, we proposed a Face 
Image Synthesis Network (FISN) ${{\cal N}_{{\rm{et}}}}$ , 
which was based on Pix2PixHD~\cite{RN390} as a backbone. 
FISN receives ${{\mathbf{I}}_{\mathbf{fac}}}\in {{\mathbb{R}}^{H\times W\times 3}}$  and ${\mathbf{M}_{\boldsymbol{\alpha }}}$  as inputs.
The detailed architecture is shown in Fig.\ref{fig:overall}.
To fuse ${{\mathbf{I}}_{\mathbf{fac}}}$  and ${\mathbf{M}_{\boldsymbol{\alpha} }}$ , 
we used Spatial Feature Transform 
(SFT) layer~\cite{RN327} learned a mapping function ${\cal M}$
that outputs a parameter pair  $\left( {\gamma ,\beta } \right) $ 
based on the prior condition ${\Psi}$ from the features ${\mathbf{M}_{\boldsymbol{\alpha}} }$. 
A pair of affine transformation parameters $\left( {\gamma ,\beta } \right)$ 
model the prior ${\Psi}$ . Here, the mapping equation 
can be expressed as $(\gamma ,\beta ){{ }} = M\left( \Psi  \right).$  After obtaining $\left( {\gamma ,\beta } \right)$ , 
the transformation is carried out by the SFT layer:
\begin{equation}
SFT({\mathbf{F}_{{\mathbf{map}}}}|\gamma ,\beta ){{ }} = \gamma  \odot F + \beta 
\end{equation}
where ${{\mathbf{F}}_{\mathbf{map}}}$  denotes the feature maps from $\mathbf{{{I}}_{fac}}$,$\odot$ denotes Hadamard product.
\\
Therefore, we conditioned spatial information ${\mathbf{M}_{\boldsymbol{\alpha}} }$  
on style data ${{\mathbf{I}}_{\mathbf{fac}}}$  and generated affine parameters   
$({{{x}}_{{i}}}{{,}}{{{y}}_i})$ followed $({{{x}}_{{i}}}{{,}}{{{y}}_i}){{ = }}{{\cal N}_{{{et}}}}\left( {{{{I}}_{{{fac}}}},{M_\alpha }} \right)$ . 
Related research~\cite{RN327} showed that ordinary 
normalization layers would "wash away" semantic information. 
To transfer  $({{{x}}_{{i}}}{{,}}{{{y}}_i})$ to new mask input ${\mathbf{M}_{\boldsymbol{\gamma}}}$ , we utilized semantic 
region-adaptive normalization (SEAN)~\cite{RN316} on residual 
blocks ${{{z}}_{{i}}}$ in the FISN. Let $H$, $W$ and $C$ be the height, width 
and the number of channels in the activation map of the 
deep convolutional network for a batch of $N$ samples. The 
modulated activation value at the site was defined as:
\begin{equation}
    SEAN\left( {{z_i},{x_i},{y_i}} \right) = {x_{{i}}}\frac{{{z_i} - \mu \left( {{z_i}} \right)}}{{\sigma \left( {{z_i}} \right)}} + {y_i}
\end{equation}
where $\mu \left( {{z_i}} \right)$  and $\sigma \left( {{z_i}} \right)$ are the mean and standard deviation of the activation $\left( {{{n}} \in N,c \in C,y \in H,x \in W} \right)$ in channel c :
\begin{equation}
    \mu \left( {{z_i}} \right){\rm{ = }}\frac{1}{{NHW}}\sum\limits_{n,y,x} {{h_{n,c,y,x}}}
\end{equation}
\begin{equation}
    \sigma \left( {{z}_{i}} \right){=}\sqrt{\frac{1}{NHW}\sum\limits_{n,y,x}{\left( {{(h_{n,c,y,x}^{{}})}^{2}}-\mu {{\left( {{z}_{i}} \right)}^{2}} \right)}}
\end{equation}
FISN is a generator that learns the style 
mapping between ${{\mathbf{I}}_{\mathbf{fac}}}$ and ${\mathbf{M}_{\boldsymbol{\gamma}}}$ according to the 
spatial information provided by ${\mathbf{M}_{\boldsymbol{\alpha}} }$. 
Therefore, face features (\textit{e.g.} eyes style) in ${{\mathbf{I}}_{\mathbf{fac}}}$  
are shifted to the corresponding position on  ${\mathbf{M}_{\boldsymbol{\gamma}} }$
so that FISN can synthesis image ${{\mathbf{I}}_{{\mathbf{out}}}}$ which removed occlusion.
\subsubsection{Loss Function}
The design of our loss function for FISN is inspired by 
Pix2PixHD~\cite{RN390}, MaskGAN~\cite{RN311} and SEAN~\cite{RN316}, 
which contains three components:
\\$(1)\ Adversarial\ loss$. Let ${{\text{D}}_{\text{1}}}$ and ${{\text{D}}_{\text{2}}}$ be two discriminators at different scales,  
${{\cal L}_{GAN}}$ is the conditional adversarial loss defined by
\begin{equation}
{{\mathcal{L}}_{GAN}}{=}\mathbb{E}\left[ \log \left( {\text{D}_{1,2}}\left( {\mathbf{I}_{\mathbf{fac}}},{\mathbf{M}_{{\boldsymbol{\alpha}} }} \right) \right) \right]+\mathbb{E}\left[ 1-\log \left( {\text{D}_{1,2}}\left( {\mathbf{I}_{\mathbf{out}}},{{\mathbf{M}}_{\boldsymbol{\alpha} }} \right) \right) \right]
\end{equation}
$(2)\ Feature\ matching\ loss$~\cite{RN390}. Let $T$ be the total number of layers in discriminator $\text{D}$  .${{\mathcal{L}}_{\text{f}ea}}$ is the feature matching loss which computed the $\text{L}_{1}$ distance between the real and generated face image defined by
\begin{equation}
    {{\mathcal{L}}_{\text{fe}a}}{=}\mathbb{E}\sum\limits_{i=1}^{T}{{{\left\| D_{1,2}^{(i)}\left( {\mathbf{I}_{\mathbf{fac}}},{{\mathbf{M}}_{\boldsymbol{\alpha} }} \right)-D_{1,2}^{(i)}\left( {\mathbf{I}_{\mathbf{out}}},{\mathbf{M}_{\mathbf{\alpha} }} \right) \right\|}_{1}}}
\end{equation}
$(3)\ Perceptual\ loss$~\cite{RN757}. Let $\text{N}$ be the total 
number of layers used to calculate the perceptual loss,
${{F}^{(i)}}$  be the output feature maps of the $i$th layer 
of the VGG network~\cite{RN617}.${{\cal L}_{{{per}}}}$ is the 
perceptual loss which computes 
the ${{L}_{1}}$ distance between the real and generated face image defined by
\begin{equation}
{{\mathcal{L}}_{\text{per}}}{=}\mathbb{E}\sum\limits_{i=1}^{N}{\frac{1}{{{M}_{i}}}[{{\left\| {{F}^{\left( i \right)}}\left( {\mathbf{I}_{\mathbf{fac}}} \right)-{{F}^{\left( i \right)}}\left( {\mathbf{I}_{\mathbf{out}}} \right) \right\|}_{1}}]}
\end{equation}
The final loss function of FISN used in our experiment is made up of the above-mentioned three loss terms as:
\begin{equation}
    {{\mathcal{L}}_{FISN}}{=}{{\mathcal{L}}_{GAN}}+{{\lambda }_{1}}{{\mathcal{L}}_{\text{fea}}}+{{\lambda }_{2}}{{\mathcal{L}}_{\text{per}}}
\end{equation}
where we set ${{\lambda }_{1}}{=}{{\lambda }_{2}}{=10}$ respectively in our experiments. 
\subsection{Camera and Illumination Model}
Given an face image, 
we adopt  the Basel Face Model (BFM)~\cite{RN45}. 
 After the 3D face is reconstructed, it can be projected 
onto the image plane with the perspective projection:
\begin{equation}
{{V}_{2d}}\left( \mathbf{P} \right)=f*{\mathbf{P}_{\mathbf{r}}}*{\mathbf{R}}*{\mathbf{S}_{\mathbf{mod} }}+{\mathbf{t}_{\mathbf{2d}}}
\end{equation}
where ${{V}_{2d}}\left( \mathbf{P} \right)$ denotes the projection 
function 
that turned the 3D model into 2D face 
positions, $f$ denotes the scale factor, 
 ${\mathbf{P}_{\mathbf{r}}}$ denotes the projection matrix,$\mathbf{R}\in SO(3)$ denotes the rotation matrix,
 ${{\mathbf{S}}_{\mathbf{mod}}}$ denotes the shape of the face
 and ${\mathbf{t}_{\mathbf{2d}}}\in {{\mathbb{R}}^{3}}$  denotes the translation vector.
 \\
We approximated the scene illumination with Spherical Harmonics (SH)~\cite{RN239} for face. 
Thus, we can compute the face as Lambertian 
surface and skin texture follows:
\begin{equation}
\mathbf{C}\left( {\mathbf{r}_{\mathbf{i}}},{\mathbf{n}_{\mathbf{i}}},\boldsymbol{\gamma}  \right)={\mathbf{r}_{\mathbf{i}}}\odot \sum\limits_{b=1}^{{{B}^{2}}}{{\boldsymbol{\gamma }_{\mathbf{b}}}{{\Phi }_{{b}}}\left( {\mathbf{n}_{\mathbf{i}}} \right)}
\end{equation}
where ${\mathbf{r}_{\mathbf{i}}}$ denotes skin reflectance,
 ${\mathbf{n}_{\mathbf{i}}}$ denotes surface normal,  
$\odot$ denotes the Hadamard product, 
$\boldsymbol{\gamma} \in {{\mathbb{R}}^{9}}$ under monochromatic 
lights condition, ${{\Phi }_{b}}:{{\mathbb{R}}^{3}}\to \mathbb{R}$  denotes SH basis function, 
$B$ denotes the number of spherical harmonics bands and ${\boldsymbol{\gamma }_{\mathbf{b}}}\in {{\mathbb{R}}^{3}}$  
(here we set $B=3$ ) denotes the corresponding SH coefficients.
\\
Therefore, parameters to be learned can be 
denoted by a vector
 $\boldsymbol{y}=(\boldsymbol{\widetilde{{{\alpha }_{i}}},\widetilde{{{\beta }_{i}}},\gamma ,p})\in {{\mathbb{R}}^{175}}$,
where  $\mathbf{p}\in {{\mathbb{R}}^{6}}=\{\boldsymbol{pitch,yaw,roll,f,{{t}_{2D}}\}}$ denotes face poses. 
In this work, we used a fixed ResNet-50~\cite{RN662} network 
to regress these coefficients. The loss function 
of ResNet-50 follows Eq.\ref{3Dzonggongshi}.
We then got the fundamental shape  ${\mathbf{S}_{\mathbf{base}}}$
(coordinate,\textit{e.g.} $x,y,z$)  and the coarse texture ${\mathbf{T}_{\mathbf{coa}}}$ 
(albedo,\textit{e.g.} $r,g,b$). We used a coarse-to-ﬁne 
network based on graph convolutional 
networks of Lin \textit{et al.}~\cite{RN150} for producing the fine 
texture ${\mathbf{T}_{\mathbf{fin}}}$.
\subsection{Loss Function of 3D Reconstruction}
Given a generated image ${{\mathbf{I}}_{\mathbf{out}}}$ ,we used the ResNet to regress the 
corresponding coefficient ${y}$. 
The design of loss function for ResNet contained 
four components:
\\(1) Landmark Loss. 
As facial landmarks convey the structural 
information of the human face, we used landmark 
loss to measure how close projected shape landmark 
vertices to the corresponding landmarks in the 
image ${{\mathbf{I}}_{\mathbf{out}}}$.
We ran the landmark prediction module ${{\mathcal{N}}_{lmk}}$ to 
detect $68$ 
landmarks $\left\{ {z}_{lmk}^{\left( n \right)} \right\}$ from the 
training images. We obtained landmarks $\left\{ l_{y}^{\left( n \right)} \right\}$  
from rendering facial images. Then, we computed the loss as:
\begin{equation}
    {{\mathcal{L}}_{{lmk}}}\left( y \right)\text{=}\frac{1}{N}\sum\limits_{\text{n}=1}^{N}{\left\| \text{z}_{\text{lmk}}^{\left( n \right)}-l_{y}^{(n)} \right\|_{2}^{2}}
\end{equation}
where ${{\left\| \cdot  \right\|}_{2}}$ denotes the ${{L}_{2}}$ norm.
\\
(2) Accurate Pixel-wise Loss.
The rendering layer renders back an image $\mathbf{I}_{\mathbf{y}}^{^{(i)}}$  
to compare with the image $\mathbf{I}_{\mathbf{out}}^{(i)}$.
The pixel-wise loss is formulated as:
\begin{equation}
{{\mathcal{L}}_{\text{pix}}}\left( y \right)\text{=}\frac{\sum\nolimits_{i\in \mathcal{M}}{{{P}_{i}}\cdot }{{\left\| \mathbf{I}_{\mathbf{out}}^{(i)}-{\mathbf{I}}_{\mathbf{y}}^{(i)} \right\|}_{2}}}{\sum\nolimits_{i\in \mathcal{M}}{{{P}_{i}}}}
\end{equation}
where $i$ denotes pixel index,  
$\mathcal{M}$ is the reprojected face region which obtained 
with landmarks~\cite{RN310}, ${{\left\| \cdot  \right\|}_{2}}$ denotes 
the ${{L}_{{2}}}$ norm and ${{P}_{i}}$  
is occlusion attention coefficient which is 
described as follows.
To gain robustness to accurate texture, we 
set $P_{i}=\begin{cases}1 \;\;\; \text{if} \;\; i\in \text{facial features of}\;{{M}_{\alpha }}\\0.1 \;\; \text{otherwise}\end{cases}$
    for each pixel $i$.
\\
(3) Regularization Loss. To prevent shape deformation and texture degeneration, we introduce the prior distribution to the parameters of the face model. We add the regularization loss as:
\begin{equation}
    {{\mathcal{L}}_{\text{reg}}}\text{=}{{\omega }_{\alpha }}{{\left\| \widetilde{{{\boldsymbol{\alpha }}_{\mathbf{i}}}} \right\|}^{2}}+{{\omega }_{\beta }}{{\left\| \widetilde{{{\boldsymbol{\beta }}_{\mathbf{i}}}} \right\|}^{2}}
\end{equation}
here, we set ${{\omega }_{\alpha }}\text{=}1.0$,${{\omega }_{\beta }}=1.\text{75e-3}$
respectively.
\\
(4)Face Features Level Loss.
To reduce the difference between 3D face with 2D image, 
we define the loss at face recognition level. 
The loss computes the feature difference 
between the input image ${{\mathbf{I}}_{\mathbf{out}}}$  and 
rendered image ${{\mathbf{I}}_{\mathbf{y}}}$  . 
We define the loss as a cosine distance:
\begin{equation}
    {{\mathcal{L}}_{ff}}\text{=1-}\frac{<G({{\mathbf{I}}_{\mathbf{out}}}),G({{\mathbf{I}}_{\mathbf{y}}})>}{\left\| G({{\mathbf{I}}_{\mathbf{out}}}) \right\|\cdot \left\| G({{\mathbf{I}}_{\mathbf{y}}}) \right\|}
\end{equation}
where $G(\cdot)$  denotes the feature extraction 
function by FaceNet~\cite{RN709},$<\cdot,\cdot>$  
denotes the inner product.
\\
In summary, the final loss function 
of 3D face reconstruction used 
in our experiment is made up of 
the above-mentioned four loss terms as:
\begin{equation}\label{3Dzonggongshi}
{{\mathcal{L}}_{3D}}{=}{{\lambda }_{3}}{{\mathcal{L}}_{lmk}}+{{\lambda }_{4}}{{\mathcal{L}}_{{pix}}}+{{\lambda }_{5}}{{\mathcal{L}}_{reg}}+{{\lambda }_{6}}{{\mathcal{L}}_{{ff}}}
\end{equation}
where we set ${{\lambda }_{3}}{=1}{.6e-3,}{{\lambda }_{4}}{=1}{.4,}{{\lambda }_{5}}{=3}\text{.7e-4,}{{\lambda }_{6}}{=0}{.2}$  
respectively in all our experiments.
\section{Implementation Details}
Considering 
the question of landmark predictor, 
the $300$-W dataset~\cite{RN743} has labeled 
ground truth landmarks, while the 
CelebA-HQ dataset~\cite{RN758} does not. 
We generated the ground truth of CelebA-HQ 
by the Faceboxes predictor~\cite{RN759} as the reference. 
In experiments shown in this work, we use the $256\times 256$ 
images for training the landmark predictor ${{\mathcal{N}}_{lmk}}$   
and the batch size$=16$ . The learning rate  of  ${{\mathcal{N}}_{lmk}}$ is $10e-4$. 
We use the trained face parsing model ${{\mathcal{N}}_{\alpha }}$~\cite{RN311} 
to generate ${{\mathbf{M}}_{\boldsymbol{\alpha }}}$. We obtain ${{\mathbf{M}}_{\boldsymbol{\gamma }}}$ 
according to Algorithm \ref{suanfa:001}. 
FISN follows the design of Pix2PixHD~\cite{RN390} with 
four residual blocks. To train the FISN, we 
used the CelebAMask-HQ
dataset which has 
$30000$ semantic labels with a size of $512\times 512$. 
Each label clearly marked the facial features of the face.
\\
FISN does not use any ordinary normalization 
layers (\textit{e.g.} Instance Normalization) which will 
wash away style information. Before training the 
ResNet, we take the weights from
pre-trained of R-Net~\cite{RN239} 
as initialization. We set the input image size to 
$224\times 224$ and the number of vertices to $35709$. 
We design our texture refinement network based on 
the Graph Convolutional Network method of 
Lin \textit{et al.}~\cite{RN150}. We do not adopt any fully-connected 
layers or convolutional layers in the refinement network 
refer to related research~\cite{RN150}.This will reduce 
the performance of the module.
\section{Experimental Results}
\subsection{Qualitative Comparisons with Recent Works} 
\begin{figure}[htb]
    \centering
\includegraphics[width=0.80\textwidth]{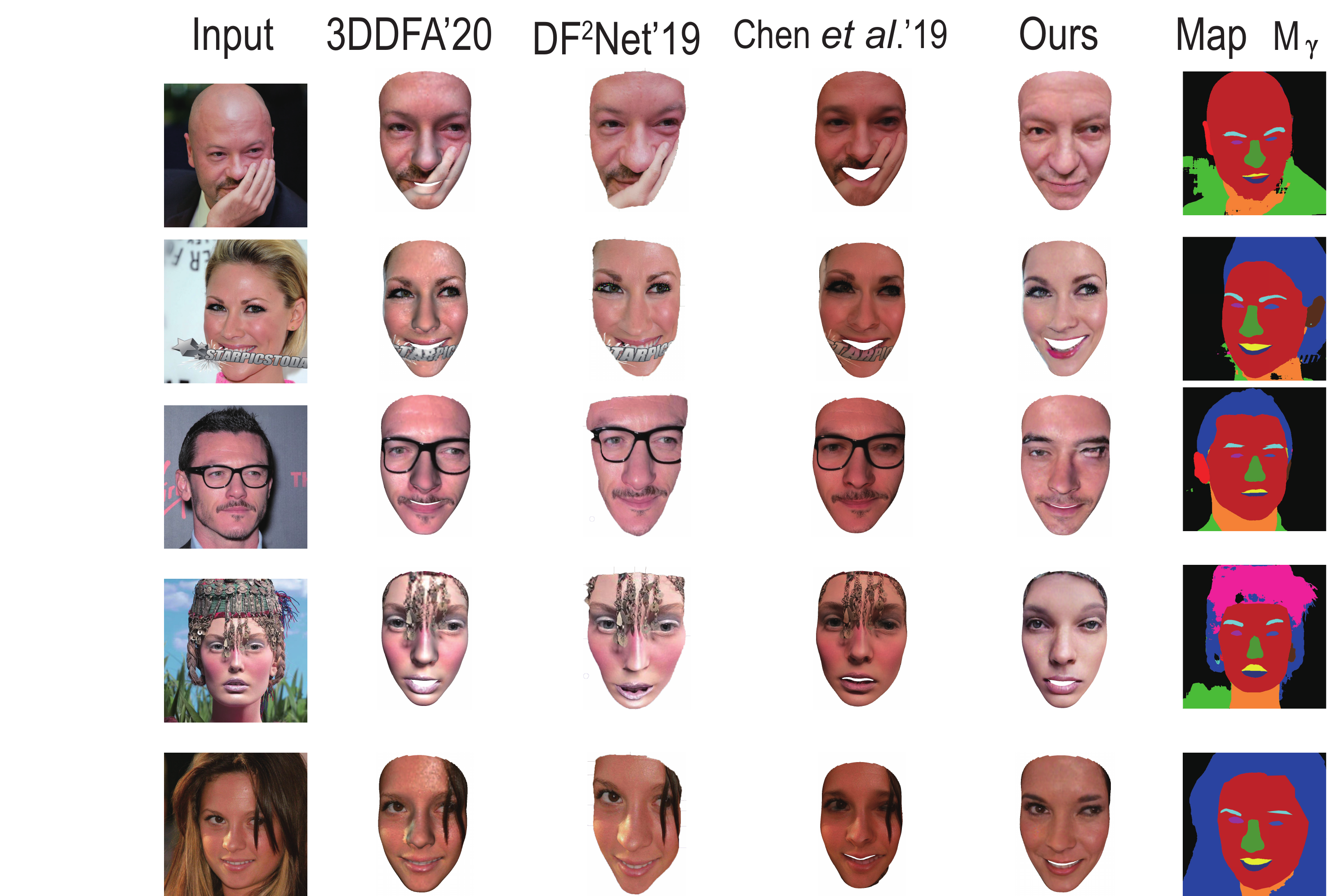}
\caption{Comparison of qualitative results. 
Baseline methods from left to right: 
3DDFA, 
$\text{D}{{\text{F}}^{\text{2}}}\text{Net}$, 
Chen \textit{et al.} 
and our method.} \label{bijiaotu}
\end{figure}

Fig.\ref{bijiaotu} shows our results compared 
with the other work. The last two columns 
show our results.The remaining columns 
demonstrate the results of 3DDFA~\cite{RN186},
$\text{D}{{\text{F}}^{\text{2}}}\text{Net}$  
~\cite{RN256} and  Chen \textit{et al.}~\cite{RN251}. Qualitative 
results show that our method surpasses 
other methods. Fig.\ref{bijiaotu} shows that our method 
can reconstruct a complete face model under 
occlusion scenes such as glasses, jewelry, 
palms, and hair. 
Other methods focused on 
generating high-resolution face textures.
These frameworks cannot effectively deal with occluded scenes.
\subsection{Quantitative Comparison} 
\subsubsection{Comparison result on the MICC Florence datasets}
\begin{figure}[htb]
    \centering
\includegraphics[width=0.39\textwidth]{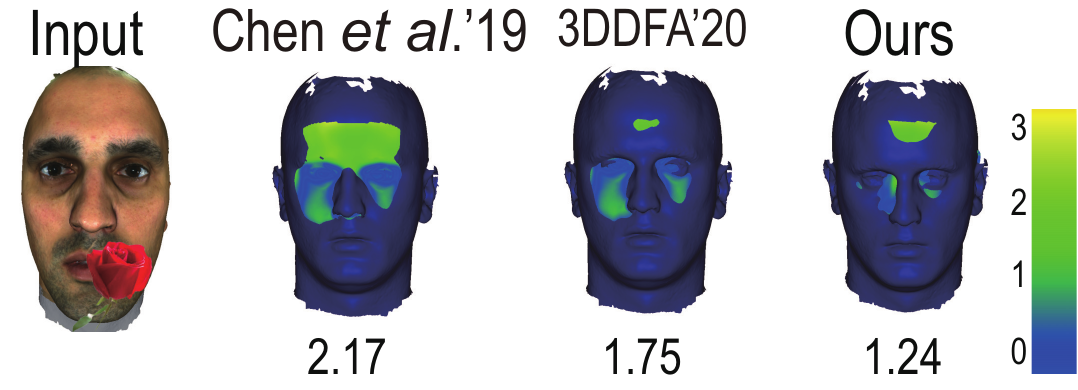}
\caption{Comparison of error heat maps on the MICC Florence datasets.
Digits denote $90\%$ error (mm).} \label{chayitu}
\end{figure} 
MICC Florence dataset~\cite{bagdanov2011florence} is a 3D face 
dataset that contains $53$ faces with their 
ground truth models. 
We artificially added some occluders as input. 
We calculated the average $90\%$ largest 
error between the generative model and the 
ground truth model.Fig.\ref{chayitu} shows that 
our method can effectively handle occlusion.\\
\begin{figure}[htb]
    \centering
\includegraphics[width=0.50\textwidth]{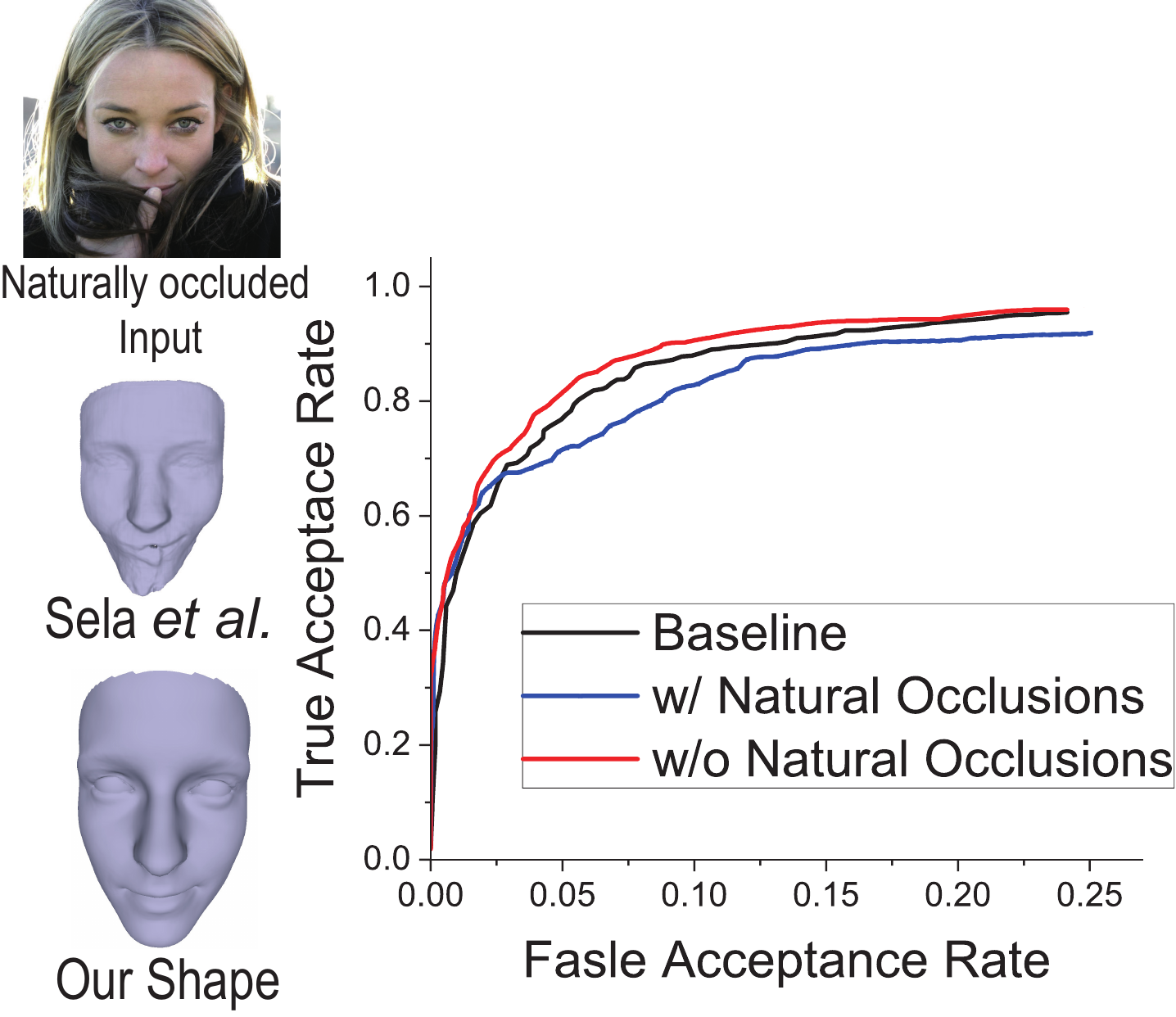}
\caption{ Reconstructions with occlusions. 
Left: Qualitative results of Sela \textit{et al.}~\cite{RN200} 
and our shape. Right: LFW veriﬁcation 
ROC for the shapes, with and without 
occlusions.} \label{origin_draw}
\end{figure}
\subsubsection{Occlusion invariance of the foundation shape}
Our choice of using the ResNet-50 to regress the shape 
coefficients is motivated by the unique robustness 
to extreme viewing conditions in the paper of 
Deng \textit{et al.}~\cite{RN239}.To fully support the application 
of our method to occluded face images, we test 
our system on the Labeled Faces in the Wild 
datasets (LFW)~\cite{RN764}. We used the same face test 
system from Anh \textit{et al.}~\cite{RN226}, and we refer 
to that paper for more details.
\\
Fig. \ref{origin_draw} (left) shows the sensitivity 
of the method of Sela et al.~\cite{RN200}. Their result 
clearly shows the outline of a finger. 
Their failure may be due to more focus on 
local details, which weakly regularizes 
the global shape. However, our method 
recognizes and regenerates the occluded area. 
Our method much robust provides a natural 
face shape under common occlusion scenes. Though 
3DMM also limits the details of shape, 
we use it only as a foundation and add 
refined texture separately.
\begin{table}
    \centering
    \caption{Quantitative evaluations on LFW.}
    \label{biaoge:01}
    \begin{tabular}{p{0.9in}c c c c}
    \hline
    \multicolumn{1}{l}{Method} & \multicolumn{1}{l}{100\%-EER} & \multicolumn{1}{l}{Accuracy} & nAUC       \\ \hline
    Tran \textit{et al.}~\cite{RN41}\qquad  & $89.40\pm1.52$\qquad\qquad                    & $89.36\pm1.25$\qquad\qquad                   & $95.90\pm0.95$ \\
    Ours (w/ Occ)\qquad                            & $85.75\pm1.12$\qquad\qquad                    & $86.49\pm0.97$\qquad\qquad                   & $93.89\pm1.31$ \\
    Ours (w/o Occ)\qquad                           & $90.57\pm1.43$\qquad\qquad                    & $89.87\pm0.71$\qquad\qquad                   & $96.59\pm0.37$ \\ \hline
    \end{tabular}
\end{table}
\\
We further quantitatively 
verify the robustness of our 
method to occlusions. 
Table \ref{biaoge:01} (top) reports veriﬁcation results on the 
LFW benchmark with and without occlusions 
(see also ROC in Fig.\ref{origin_draw} (right)). Though 
occlusions clearly impact recognition, 
this drop of the curve is limited, 
demonstrating the robustness of our method.
\section{Conclusions}
In this work, we present a novel single-image 
3D face reconstruction method under occluded scenes 
with high ﬁdelity textures. Comprehensive experiments have shown 
that our method outperforms previous methods by a 
large margin in terms of both accuracy and robustness. 
Future work includes combining our method with Transformer 
architecture to further improve accuracy.
\section{Acknowledgment}
This paper is supported by National Natural Science Foundation of China (No. 62072020), National Key Research and Development Program of China (No. 2017YFB1002602), Key-Area Research and Development Program of Guangdong Province (No. 2019B010150001) and the Leading Talents in Innovation and Entrepreneurship of Qingdao (19-3-2-21-zhc).

\bibliographystyle{splncs04}
\bibliography{ref.bib}  

\printindex

\end{document}